\documentclass{article}

    \PassOptionsToPackage{numbers, compress}{natbib}
    \usepackage[preprint]{neurips_2019}

\usepackage[utf8]{inputenc} % allow utf-8 input
\usepackage[T1]{fontenc}    % use 8-bit T1 fonts
\usepackage{hyperref}       % hyperlinks
\usepackage{url}            % simple URL typesetting
\usepackage{booktabs}       % professional-quality tables
\usepackage{amsfonts}       % blackboard math symbols
\usepackage{nicefrac}       % compact symbols for 1/2, etc.
\usepackage{microtype}      % microtypography
\usepackage{xcolor}
\usepackage{graphicx}
\usepackage{amsmath}
\usepackage{amssymb}
\usepackage{multirow}
\usepackage{wrapfig}
\usepackage{color}
\usepackage{hyperref}
\usepackage{float} 
\usepackage{subfigure}
\usepackage{titlesec}
\usepackage{authblk}
\hypersetup{
    colorlinks = true,
    citecolor = green
}

\DeclareMathOperator{\birth}{birth}
\DeclareMathOperator{\death}{death}
\DeclareMathOperator{\dgm}{Dgm}
\usepackage{url}
\usepackage{graphicx} % more modern
\usepackage{algorithm}
\usepackage{algorithmicx,algpseudocode}
\usepackage{multirow}
\usepackage{amsmath}
\usepackage{amssymb}
\usepackage{amsthm}
\usepackage{bbm}
\usepackage{xspace}
\usepackage{enumitem}
\usepackage{color}
\usepackage{placeins} % \FloatBarrier command creates a barrier that floats cannot cross

\usepackage{dsfont}

\usepackage{comment}
\usepackage{color}
\usepackage{bm}
\usepackage{array}
\usepackage{wrapfig}
\usepackage{lipsum}
\usepackage{enumitem}
\usepackage{relsize}
\setenumerate{leftmargin=*}
\setenumerate[1]{label=(\arabic*),align=left}
\setenumerate[2]{label=(\alph*)}

\newtheorem{theorem}{Theorem}
\makeatletter
\DeclareRobustCommand\onedot{\futurelet\@let@token\@onedot}
\def\@onedot{\ifx\@let@token.\else.\null\fi\xspace}
\makeatother

\def\eg{{e.g}\onedot} 
\def\ie{{i.e}\onedot}

\def\wrt{w.r.t\onedot} 

\def\etal{\emph{et al}\onedot}

\numberwithin{equation}{section}
\newcommand{\myparagraph}[1]{\noindent\textbf{#1}}

\title{Topology-Preserving Deep Image Segmentation}

\author[1]{\thanks{Correspondence to: Xiaoling Hu <xiaolhu@cs.stonybrook.edu>.}\ \ Xiaoling Hu}
\author[2]{Li Fuxin}
\author[1]{Dimitris Samaras}
\author[1]{Chao Chen}

\affil[1]{Stony Brook University} \affil[2]{Oregon State University}

\begin{document}

\maketitle

\begin{abstract}
Segmentation algorithms are prone to make topological errors on fine-scale structures, e.g., broken connections.
We propose a novel method that learns to segment with correct topology.
In particular, we design a continuous-valued loss function that enforces a segmentation to have the same topology as the ground truth, i.e., having the same Betti number.
The proposed topology-preserving loss function is differentiable and we incorporate it into end-to-end training of a deep neural network. Our method achieves much better performance on the Betti number error, which directly accounts for the topological correctness. It also performs superiorly on other topology-relevant metrics, e.g., the Adjusted Rand Index and the Variation of Information. We illustrate the effectiveness of the proposed method on a broad spectrum of natural and biomedical datasets. %, such as CREMI, ISBI12, ISBI13, DRIVE, Road, and CrackTree. 
\end{abstract}

% \setlength{\abovedisplayskip}{-1pt}
% \setlength{\belowdisplayskip}{-5pt}
%\titlespacing{\section}{0pt}{*0}{*0}
%\titlespacing{\subsection}{0pt}{*0}{*0}

\vspace{-5pt}
\section{Introduction}
\begin{figure*}[t]
\centering 
\subfigure[]{
\includegraphics[width=0.18\textwidth]{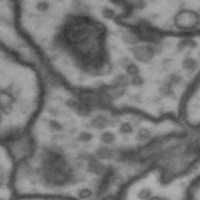}}
% \vspace{-10pt}
\subfigure[]{
\includegraphics[width=0.18\textwidth]{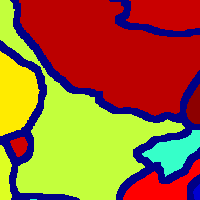}}
\subfigure[]{
\includegraphics[width=0.18\textwidth]{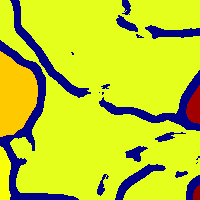}}
\subfigure[]{
\includegraphics[width=0.18\textwidth]{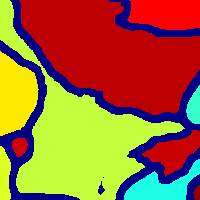}}
\vspace{-8pt}
\caption{ Illustration of the importance of topological correctness in a neuron image segmentation task. The goal of this task is to segment membranes which partition the image into regions corresponding to neurons. \textbf{(a)} an input neuron image. \textbf{(b)} ground truth segmentation of the membranes (dark blue) and the result neuron regions. \textbf{(c)} result of a baseline method without topological guarantee~\cite{fakhry2016deep}. Small pixel-wise errors lead to broken membranes, resulting in merging of many neurons into one. \textbf{(d)} Our method produces the correct topology and the correct partitioning of neurons.}
\label{fig:teaser}
\vspace{-.15in}
\end{figure*}

\vspace{-5pt}
Image segmentation, \ie, assigning  labels to all pixels of an input image, is crucial in many computer vision tasks. 
% Using various deep neural network architectures~\cite{krizhevsky2012imagenet,simonyan2014very,he2016deep}, 
State-of-the-art deep segmentation methods~\cite{long2015fully,he2017mask,chen2014semantic,chen2018deeplab,chen2017rethinking} learn high quality feature representations through an end-to-end trained deep network and achieve satisfactory per-pixel accuracy. However, these segmentation algorithms are still prone to make errors on fine-scale structures, such as small object instances, instances with multiple connected components, and thin connections. These fine-scale structures may be crucial in analyzing the \emph{functionality} of the objects. For example, accurate extraction of thin parts such as ropes and handles is crucial in planning robot actions, \eg, dragging or grasping. In biomedical images, correct delineation of thin objects such as neuron membranes and vessels is crucial in providing accurate morphological and structural quantification of the underlying system. A broken connection or a missing component may only induce marginal per-pixel error, but can cause catastrophic functional mistakes. 
See Fig.~\ref{fig:teaser} for an example.

% The problem is that, in image segmentation/recognition tasks, usually we mainly focus on either finding more representative features or designing more complex/fancy network architectures. And pixel-level loss function (MSE,BCE) are used to train the neural networks. Unsurprisingly, these methods will achieve high performance in terms of pixel accuracy, while lose some global geometry information. 

We propose TopoNet, a novel deep segmentation method that \emph{learns to segment with correct topology}. In particular, we propose a \emph{topological loss} that enforces the segmentation results to have the same topology as the ground truth, \ie, having the same \emph{Betti number} (number of connected components and handles). A neural network trained with such loss will achieve high topological fidelity without sacrificing per-pixel accuracy. The main challenge in designing such loss is that topological information, namely, Betti numbers, are discrete values. We need a continuous-valued measurement of the topological similarity between a prediction and the ground truth; and such measurement needs to be differentiable in order to backpropagate through the network. 

To this end, we propose to use theory from computational topology \cite{edelsbrunner2010computational}, which summarizes the topological information from a continuous-valued function (in our case, the likelihood function $f$ is predicted by a neural network). Instead of acquiring the segmentation by thresholding $f$ at 0.5 and inspecting its topology, \emph{persistent homology} \cite{edelsbrunner2010computational,edelsbrunner2000topological,zomorodian2005computing} captures topological information carried by $f$ over all possible thresholds. This provides a unified, differentiable approach of measuring the topological similarity between $f$ and the ground truth, called the topological loss. %Optimizing the topological loss essentially drives these structures towards their matchings. 
We derive the gradient of the loss so that the network predicting the likelihood function can be optimized \wrt the topological loss. 

\emph{TopoNet is the first end-to-end deep segmentation network with guaranteed topological correctness.} We show that when the topological loss is decreased to zero, the segmentation is guaranteed to be topologically correct, \ie, have identical topology as the ground truth. Our method is empirically validated by comparing with state-of-the-arts on natural and biomedical datasets with fine-scale structures. It achieves superior performance on metrics that encourage structural accuracy. In particular, our method significantly outperforms others on the Betti number error which exactly measures the topological accuracy. Fig.~\ref{fig:teaser} shows a qualitative result.

Our method demonstrates how topological computation and deep learning can be mutually beneficial. While our method empowers deep neural network with topological constraints, it also can be seen as a powerful approach on topological analysis in that the observation function is now learned with a highly nonlinear deep network. This enables topology to be estimated based on a semantically informed and denoised observation function. 

%This empowers topological analysis to deal with observation functions that encode semantic meanings, e.g. categorization, which was difficult to model previously. 

\myparagraph{Related work.}
The closest method to ours is by Mosinska \etal~\cite{mosinska2018beyond}, which also proposes a topology-aware loss. Instead of actually computing and comparing the topology, their approach uses the response of selected filters from a pretrained VGG19 network to construct the loss. These filters prefer elongated shapes and thus alleviate the broken connection issue. But this method is hard to generalize to more complex settings with connections of arbitrary shapes. Furthermore, even if this method achieves zero loss, its segmentation is not guaranteed to be topologically correct.

Different ideas have been proposed to capture fine details of objects, mostly revolving around deconvolution and upsampling~\cite{long2015fully,chen2014semantic,chen2018deeplab,chen2017rethinking,noh2015learning,ronneberger2015u}.
% However the connection between VGG filters and topological invariants is unclear at best. Some topological invariants such as connectivity cannot be easily  represented by VGG filters which are localized. 
However these methods focus on the prediction accuracy of individual pixels and are intrinsically topology-agnostic.
Topological constraints, \eg, connectivity and loop-freeness, have been incorporated into variational~\cite{han2003topology,le2008self,sundaramoorthi2007global,segonne2008active} and MRF/CRF-based segmentation methods~\cite{vicente2008graph,nowozin2009global,zeng2008topology,chen2011enforcing,andres2011probabilistic,stuhmer2013tree,oswald2014generalized,estrada2014tree}. 
However, these methods focus on enforcing topological constraints in the inference stage, while the trained model is agnostic of the topological prior. 
% Such model tends to make topological mistakes in inference and adds burdens to the inference task. 
% Furthermore, we are missing very rich topological information carried by the training data. For example, in Figure \ref{fig:EMExample}(a), within a specific region, the ground truth structure has a shape $Y$ with one hole in the middle. These structural information is topological and should be fully leveraged.
In neuron image segmentation, some methods \cite{funke2017deep,turaga2009maximin} directly find an optimal partition of the image into neurons, and thus avoid segmenting membranes. These methods cannot be generalized to other problems when the object of interest may not be closed loops, \eg, vessels, cracks and roads.  

For completeness, we also refer to other existing works on extracting topological features and training kernel classifiers \cite{adams2017persistence,reininghaus2015stable,kusano2016persistence,carriere2017sliced,chen2019topological}. 
In graphics, topological similarity was used to simplify and align shapes \cite{poulenard2018topological}.
As for deep neural networks, Hofer \etal \cite{hofer2017deep} proposed a CNN-based topological classifier. This method directly extracts topological information from an input image/shape/graph as input for CNN, hence cannot generate segmentations that preserve topological priors learned from the training set.
To the best of our knowledge, no existing work uses topological information as a loss for training an end-to-end deep neural network.
\vspace{-5pt}
\section{TopoNet}
\vspace{-5pt}
Our method, TopoNet, achieves both per-pixel accuracy and topological correctness by training a deep neural network with a new topological loss, $L_{topo}(f,g)$. Here $f$ is the likelihood map predicted by the network and $g$ is the ground truth. The loss function on each training image is a weighted sum of the per-pixel cross-entropy loss, $L_{bce}$, and the topological loss:
\begin{equation}
    L(f,g)= L_{bce}(f,g)+ \lambda L_{topo}(f,g),
\label{total_loss}
\end{equation}
in which $\lambda$ controls the weight of the topological loss. We assume a binary segmentation task. Thus, there is one single likelihood function $f$, whose value ranges between 0 and 1. 
%
% \textbf{Notation.} In this paper, $x \in \mathbb{R}^{H \times W}$ denotes the $H \times W$ input image. $g \in \{0, 1\}^{H \times W}$ is the ground truth label, in which 1 indicates pixels in the curvilinear structure and 0 indicates background. $\omega$ denotes the parameters of our neural network. The output of the network is an image $\hat{g} = f(x, \omega) \in [0, 1]^{H \times W}$. Especially, we focus on binary segmentation tasks in this paper.
%

\begin{figure*}[t!]
\vskip -5pt
  \centering
  \noindent\makebox[\textwidth][c] {
    \includegraphics[width=0.6\paperwidth]{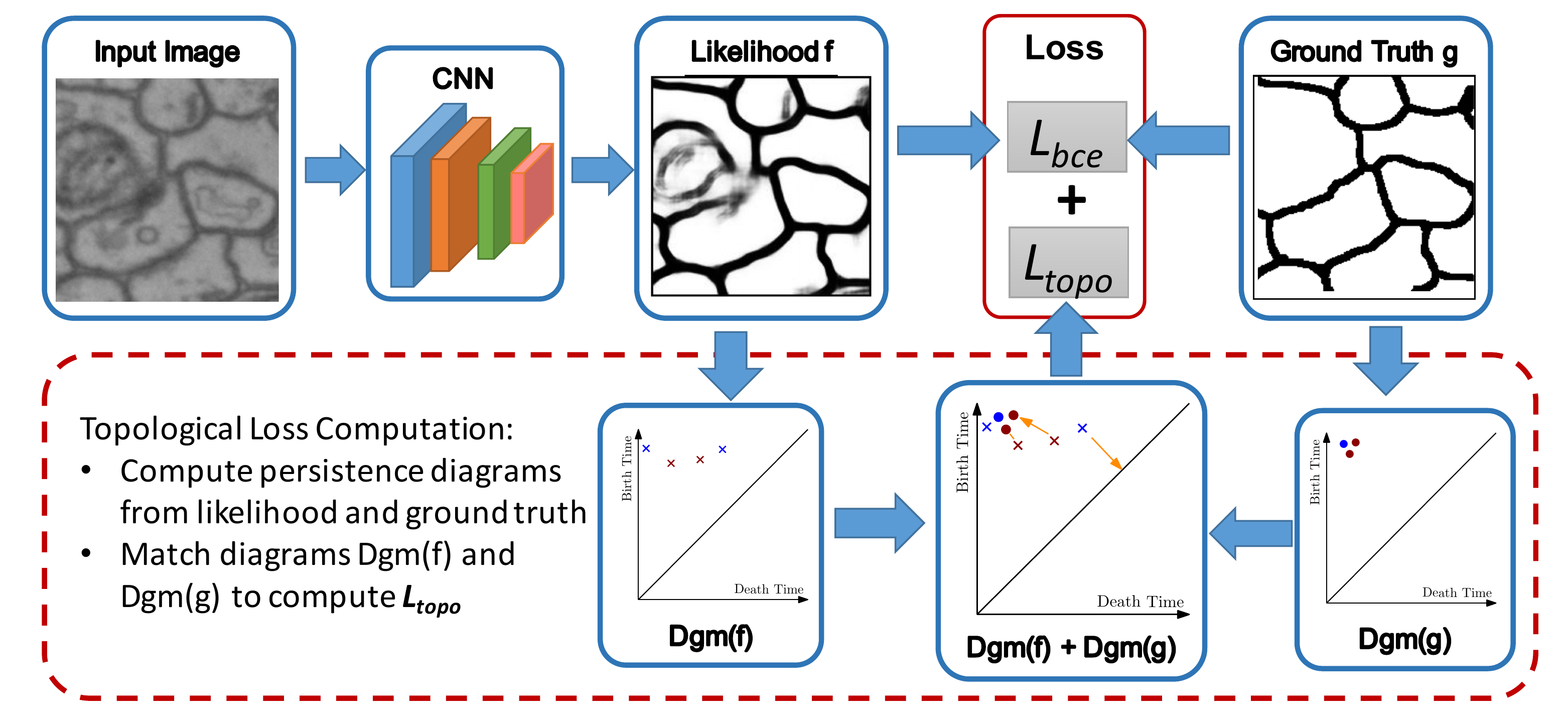}}

    \vspace{-10pt}
    \caption{An overview of our method.}
      \label{fig:architecture}
     \vskip -10pt
\end{figure*}

 In Sec.~\ref{topo_ph}, we introduce the mathematical foundation of topology and how to measure topology of a likelihood map robustly using persistent homology. In Sec.~\ref{loss_gradient}, we formalize the topological loss as the difference between persistent homology of $f$ and $g$. We derive the gradient of the loss and prove its correctness.  In Sec.~\ref{details} we explain how to incorporate the loss into the training of a neural network. Although we fix one architecture in experiments, our method is general and can use any neural network that provides pixel-wise prediction. Fig.~\ref{fig:architecture} illustrates the overview of our method.

\vspace{-5pt}
\subsection{Topology and Persistent Homology}
\label{topo_ph}
\begin{figure*}[t]
\centering 
\subfigure[]{\label{fig:2a} 
\includegraphics[width=0.18\textwidth]{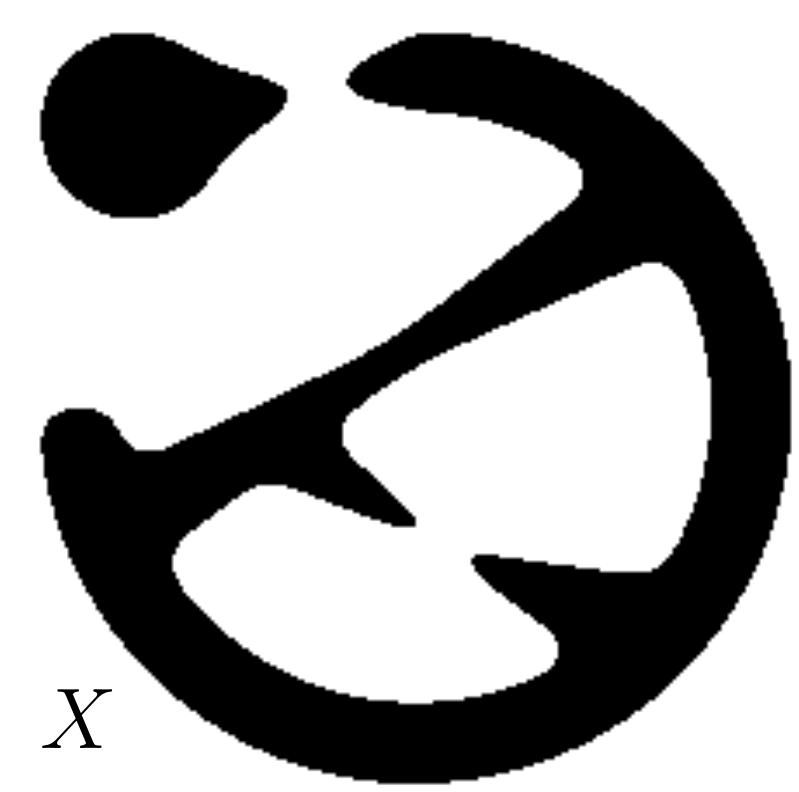}}
\subfigure[]{\label{fig:2b} 
\includegraphics[width=0.18\textwidth]{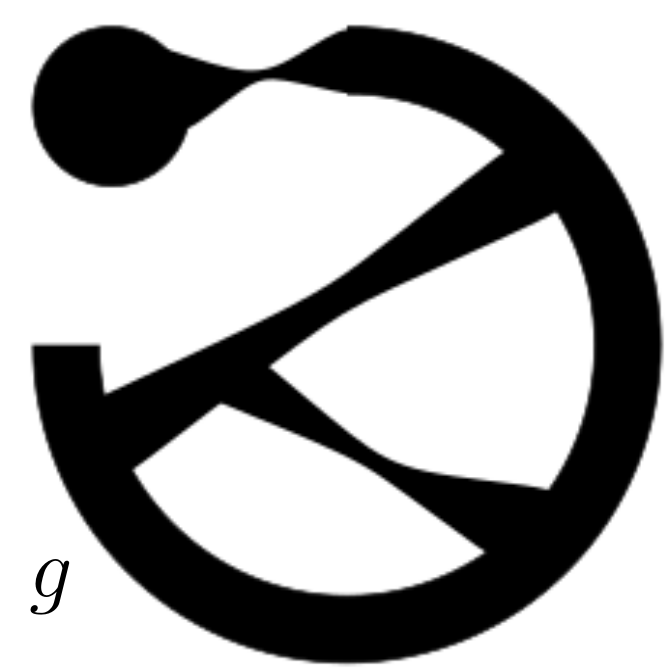}}
\subfigure[]{\label{fig:2c} 
\includegraphics[width=0.27\textwidth]{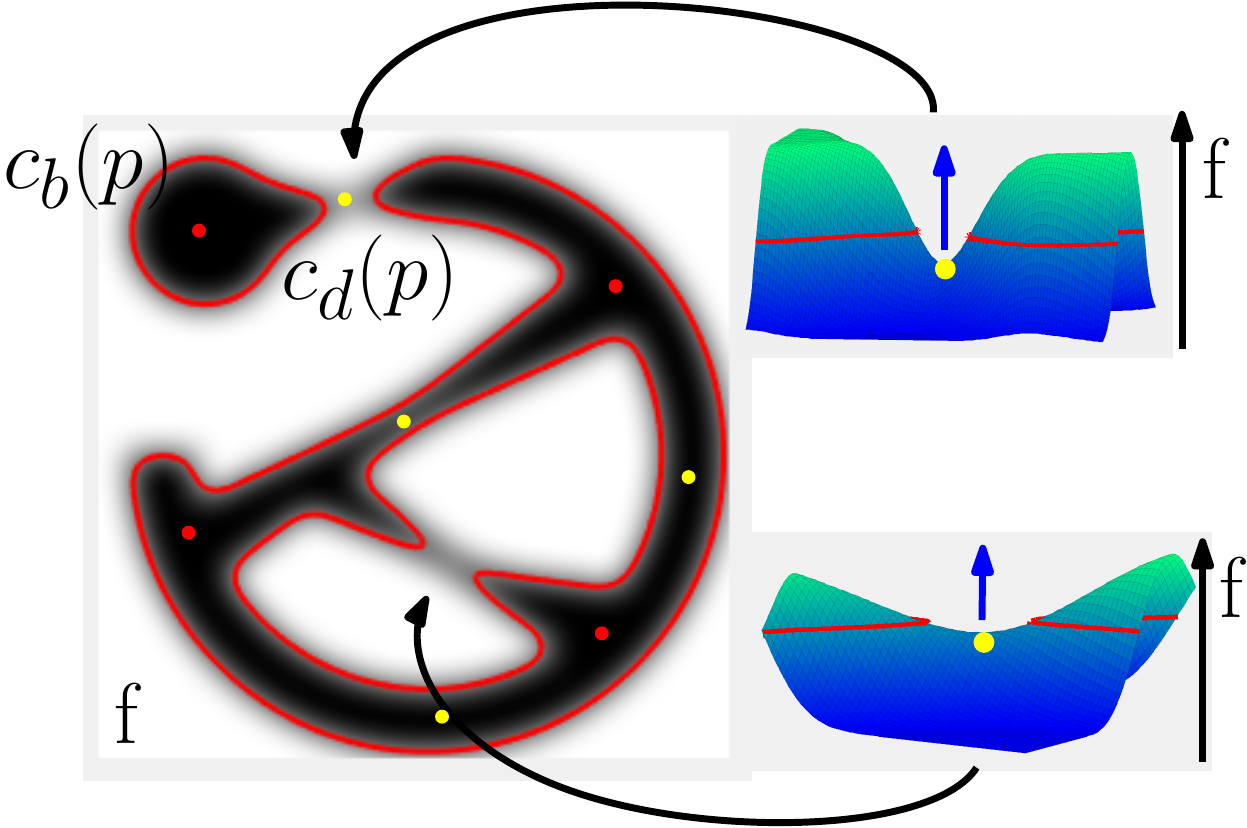}}
\subfigure[]{\label{fig:2d} 
\includegraphics[width=0.27\textwidth]{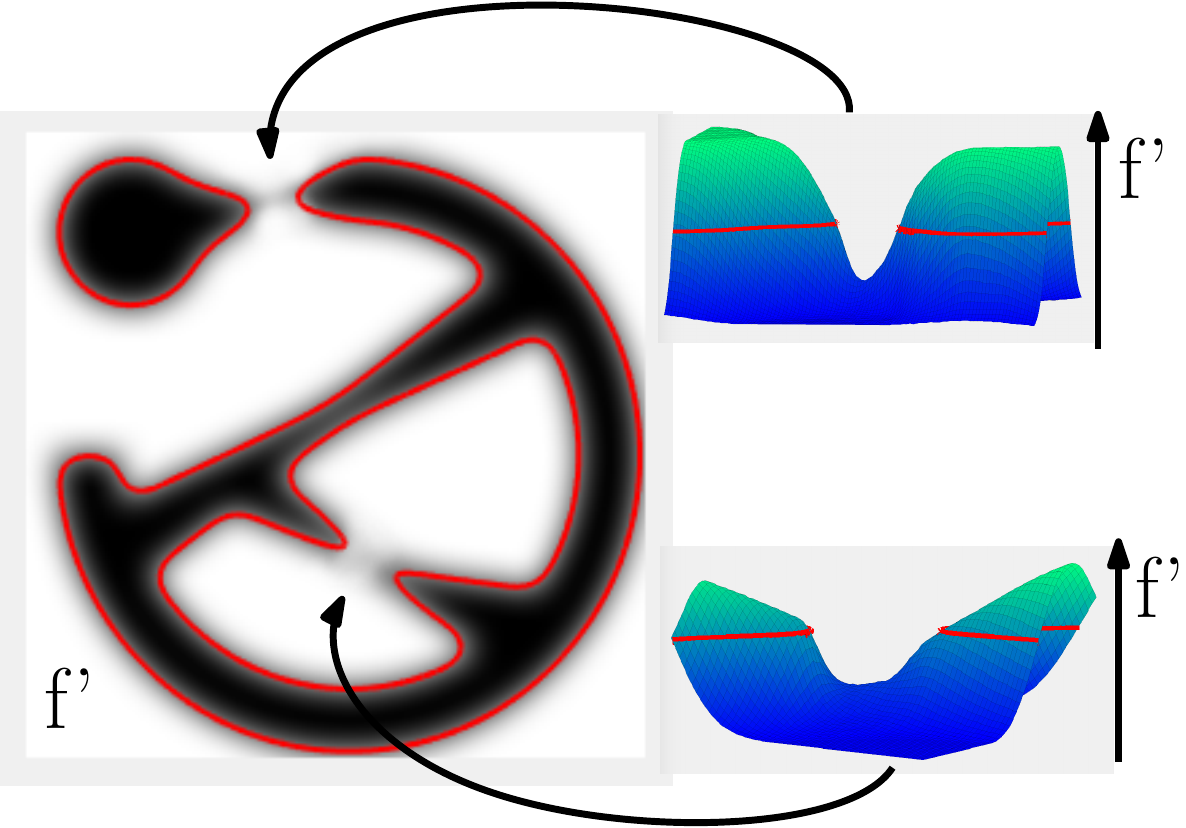}}
\vspace{-.1in}
\caption{Illustration of topology and topology of a likelihood. For visualization purposes, the higher the function values are, the darker the area is. \textbf{(a)} an example segmentation $X$ with two connected components and one handle. \textbf{(b)} The ground truth with one connected component and two handles. It can also be viewed as a binary valued function $g$. \textbf{(c)} a likelihood map $f$ whose segmentation (bounded by the red curve) is $X$. The landscape views near the broken bridge and handle are drawn. Critical points are highlighted in the segmentation.
% \ccinline{Thicken red curves. Critical points highlight.} 
\textbf{(d)} another likelihood map $f'$ with the same segmentation as $f$. But the landscape views reveal that $f'$ is worse than $f$ due to deeper gaps. }
\label{fig:similarity}
\vspace{-12pt}
\end{figure*}

\vspace{-5pt}
% In this part, we will explain how to measure the topological similarity between 2D images. 
Given a continuous image domain, $\Omega \subseteq \mathbb{R}^2$ (\eg, a 2D rectangle), we study a likelihood map $f(x, \omega): \Omega \rightarrow \mathbb{R}$, which is predicted by a deep neural network (Fig.~\ref{fig:2c}). Note that in practice, we only have samples of $f$ at all pixels. In such case, we extend $f$ to the whole image domain $\Omega$ by linear interpolation. Therefore, $f$ is piecewise-linear and is controlled by values at all pixels. A segmentation, $X \subseteq \Omega$ (Fig.~\ref{fig:2a}), is calculated by thresholding $f$ at a given value $\alpha$ (often set to $0.5$). 

Given $X$, its $d$-dimension topological structure, called a \textit{homology class}~\cite{edelsbrunner2010computational,munkres2018elements}, is an equivalence class of $d$-manifolds which can be deformed into each other within $X$.\footnote{To be exact, a homology class is an equivalent class of cycles whose difference is the boundary of a $(d+1)$-dimensional patch.} In particular, 0-dim and 1-dim structures are connected components and handles, respectively. For example, in Fig.~\ref{fig:2a}, the segmentation $X$ has two connected components and one handle. Meanwhile, the ground truth (Fig.~\ref{fig:2b}) has one connected component and two handles. Given  $X$, we can compute the number of topological structures, called the \textit{Betti number}, and compare it with the topology of the ground truth.

However, simply comparing Betti numbers of $X$ and $g$ will result in a discrete number. In order to incorporate topological prior into deep neural networks, we need to construct a continuous-valued function that can reveal subtle difference between similar structures. Fig.~\ref{fig:2c} and \ref{fig:2d} show two likelihood maps $f$ and $f'$ with identical segmentations, both different in topology w.r.t. the ground truth $g$ (Fig.~\ref{fig:2b}). However, $f$ is more preferable as we need much less effort to change it so that the thresholded segmentation $X$ is the same as $g$. In particular, look closely to Fig.~\ref{fig:2c} and \ref{fig:2d} near the broken handle and view the magnified view of the function. To restore the broken handle in Fig.~\ref{fig:2d}, we need to spend more effort to fill a much deeper gap than Fig.~\ref{fig:2c}. The same situation happens near the missing bridge between the two connected components.

% \begin{figure*}[t]
%   \centering
%   \noindent\makebox[\textwidth][c] {
%     \includegraphics[width=0.12\paperwidth]{figure/seg_annotated.pdf}
%     \includegraphics[width=0.12\paperwidth]{figure/GT.pdf}
%         \includegraphics[width=0.18\paperwidth]{figure/lh_with_seg.pdf}
%     \includegraphics[width=0.18\paperwidth]{figure/lh_bad_with_seg.pdf}}
%     \caption{(Best viewed in color) (a): an example segmentation $X$ with two connected components and one handle. (b): The ground truth with one connected component and two handles. It can be viewed as a binary valued function $g$. (c): a likelihood map f whose segmentation (bounded by red curve) is $X$. The landscape views near the broken bridge and handle are drawn. (d): another likelihood map $f'$ with almost the same segmentation (enclosed by the red curve) as $f$ . But the landscape views reveal that $f'$ is worse than $f$ due to deeper gaps. For visualization purposes, the
% higher the function values are, the darker the area is.}
%       \label{fig:similarity}
% \end{figure*}

% \subsection{Persistent Homology}
% \label{homology}

To capture such subtle structural difference between different likelihood maps, we need a holistic view. In particular, we use the theory of \emph{persistent homology}.
Instead of choosing a fixed threshold, persistent homology theory captures all possible topological structures from all thresholds, and summarize all these information in a concise format, called \textit{persistence diagram}.

All different thresholds $\alpha$ constitute a filtration, i.e., a monotonically growing sequence induced by decreasing the threshold $\alpha: \varnothing \subseteq f^{\alpha_1} \subseteq f^{\alpha_2} \subseteq ... \subseteq f^{\alpha_n} = \Omega$, where $\alpha_1 \geq \alpha_2 \geq ... \geq \alpha_n$, and $ f^{\alpha}:= \{x \in \Omega | f(x) \geq \alpha \}$. As $\alpha$ decreases, the topology of $f^{\alpha}$ changes. That means some new topological structures are born while other existing topological structures are killed. When $\alpha<\alpha_n$, only one connected component survives and never gets killed. See Fig.~\ref{fig:a} and \ref{fig:d} for example filtrations induced by the ground truth function $g$ and the likelihood function $f$.

For a continuous-valued function $f$, its \textit{persistence diagram}, $\dgm(f)$, contains a finite number of dots in 2-dimensional plane, called \textit{persistent dots}. Each persistent dot $p=(d,b) \in \dgm(f)$ corresponds to a topological structure which is born at time/threshold $b$ and killed at time/threshold $d$. Denote $\birth(p)=b$ and $\death(p)=d$.
 Fig.~\ref{fig:b} and \ref{fig:e} show the diagrams of $g$ and $f$, respectively.\footnote{Unlike traditional setting, we use the birth time as the y axis and the death time as the x axis. This is because we are using an upperstar filtration, i.e., using the superlevel set, and decreasing $\alpha$ value.} Instead of comparing discrete Betti numbers, we can use the information from persistence diagrams to compare a likelihood $f$ with the ground truth $g$ in terms of topology.

% \begin{figure*}[h]
%   \centering
%   \noindent\makebox[\textwidth][c] {
%     \includegraphics[width=0.2\paperwidth]{figure/Topo_gt.png}
%     \includegraphics[width=0.2\paperwidth]{figure/Topo_lh.png}
%     }{
%         \includegraphics[width=0.2\paperwidth]{figure/Topo_fix.png}
%     \includegraphics[width=0.22\paperwidth]{figure/Topo_dig.png}}
%     \caption{Topology analysis: From left to right, from up to bottom: ground truth, likelihood map, persistent points and persistence diagrams.}
%       \label{fig:persistent}
% \end{figure*}

% \begin{figure*}[t]
%   \centering
%   \noindent\makebox[\textwidth][c] {
%     \includegraphics[width=0.18\paperwidth]{figure/Topo_gt.png}
%     \includegraphics[width=0.18\paperwidth]{figure/Topo_lh.png}
%         \includegraphics[width=0.18\paperwidth]{figure/Topo_fix.png}
%     \includegraphics[width=0.18\paperwidth]{figure/Topo_dig.png}}
%     \caption{Topology analysis: From left to right, from up to bottom: ground truth, likelihood map, persistent points and persistence diagrams.}
%       \label{fig:persistent}
% \end{figure*}

\vspace{-5pt}
\subsection{Topological loss and the Gradient}
\label{loss_gradient}
\begin{figure*}[t]
\centering 
\subfigure[Filtration induced by the ground truth function, $g$.]{\label{fig:a} 
\includegraphics[width=0.5\textwidth]{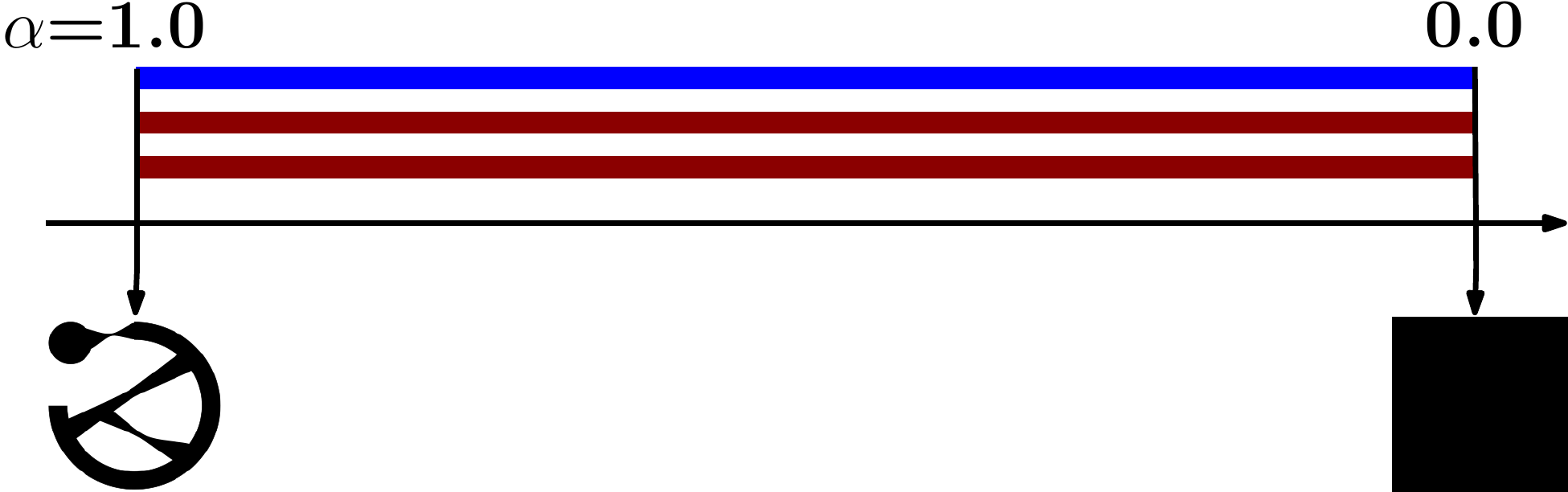}}
% \vspace{-10pt}
\hspace{0.2in}
\subfigure[$\dgm(g)$]{\label{fig:b} 
\includegraphics[width=0.15\textwidth]{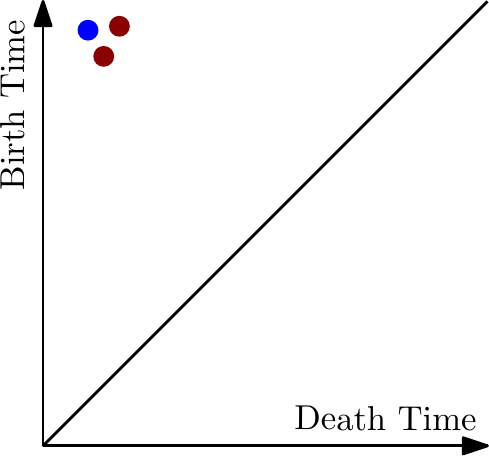}}
\hspace{0.05in}
\subfigure[$\dgm(g)$+$\dgm(f)$]{\label{fig:c} 
\includegraphics[width=0.15\textwidth]{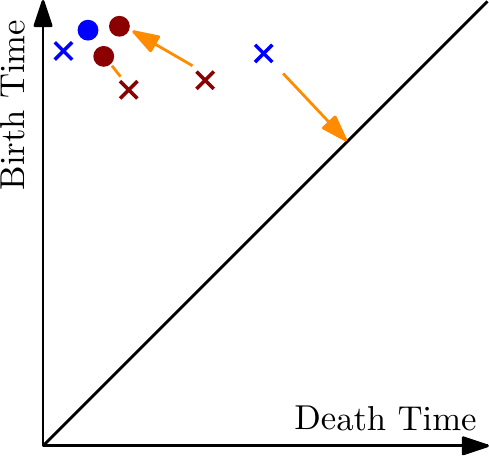}}
\subfigure[Filtration induced by the likelihood function, $f$.]{\label{fig:d} 
% \vspace{-10pt}
\includegraphics[width=0.5\textwidth]{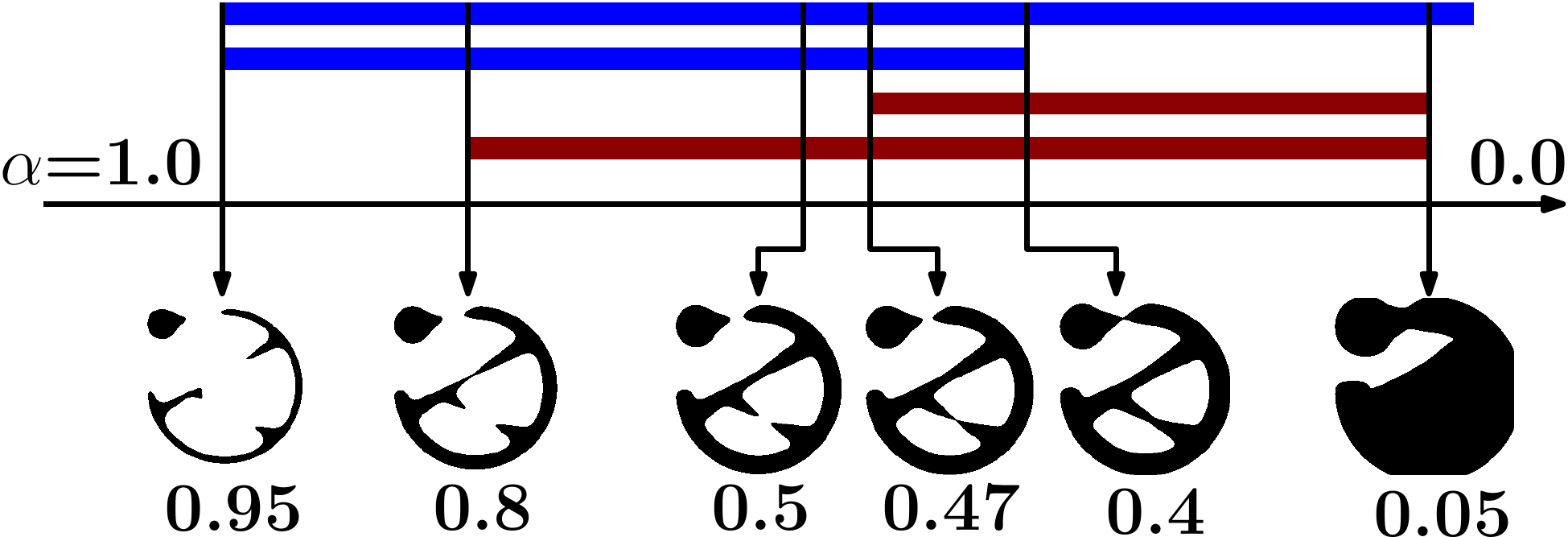}}
\hspace{0.2in}
\subfigure[$\dgm(f)$]{\label{fig:e}
% \vspace{-10pt}
\includegraphics[width=0.15\textwidth]{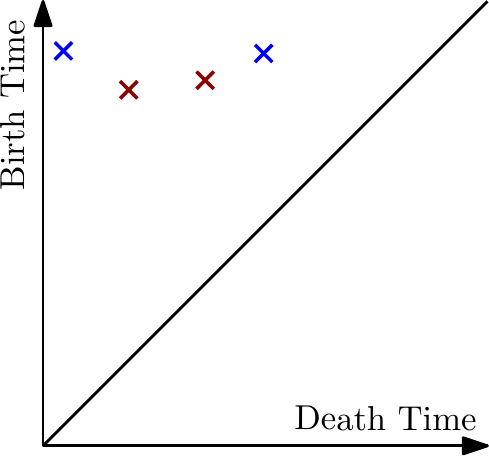}}
\hspace{0.05in}
\subfigure[$\dgm(g)$+$\dgm({f'})$]{\label{fig:f} 
% \vspace{-10pt}
\includegraphics[width=0.15\textwidth]{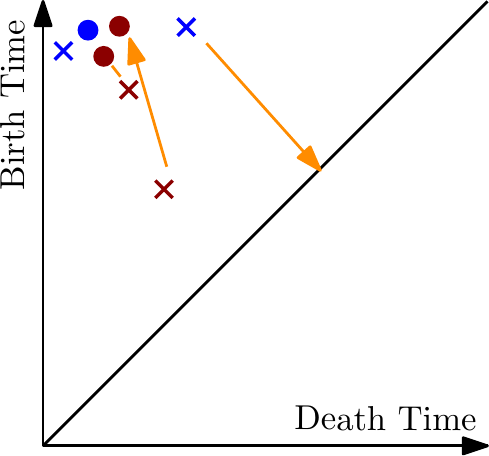}}
\vspace{-8pt}
\caption{An illustration of persistent homology. \textbf{Left} the filtrations on the ground truth function $g$ and the likelihood function $f$. The bars of blue and burgundy colors are connected components and handles respectively. \textbf{(a)} For $g$, all structures are born at $\alpha = 1.0$ and die at $\alpha = 0$. \textbf{(d)} For $f$, from left to right, birth of two components, birth of the longer handle, segmentation at $\alpha = 0.5$, birth of the shorter handle, death of the extra component, death of both handles. \textbf{(b)} and \textbf{(e)} the persistence diagrams of $g$ and $f$. \textbf{(c)} the overlay of the two diagrams. Orange arrows denote the matching between the persistent dots. The extra component (a blue cross) from the likelihood is matched to the diagonal line and will be removed if we move $\dgm(f)$ to $\dgm(g)$. \textbf{(f)} the overlay of the diagrams of $g$ and the worse likelihood $\dgm({f'})$. The matching is obviously more expensive.}
\label{fig:persistent}
\vspace{-10pt}
\end{figure*}

\vspace{-5pt}
We are now ready to formalize the topological loss, which measures the topological similarity between the likelihood $f$ and the ground truth $g$. We abuse the notation and also view $g$ as a binary valued function.
We use the dots in the persistence diagram of $f$ as they capture all possible topological structures $f$ potentially has. We slightly modified the \textit{Wasserstein distance} for persistence diagram \cite{cohen2010lipschitz}.
For persistence diagrams $\dgm(f)$ and $\dgm(g)$, we find a best one-to-one correspondence between the two sets of dots, and measure the total squared distance between them. \footnote{To be exact, the matching needs to be done on separate dimensions. Dots of 0-dim structures (blue markers in Fig.~\ref{fig:b} and \ref{fig:e}) should be matched to the diagram of 0-dim structures. Dots of 1-dim structures (red markers in Fig.~\ref{fig:b} and \ref{fig:e}) should be matched to the diagram of 1-dim  structures. }  
Let $\Gamma$ be the set of all possible bijections between $\dgm(f)$ and $\dgm(g)$.
The topological loss $L_{topo}(f,g)$ is:
% \vspace{-0pt}
\begin{equation}
% \vspace{-10pt}
\label{topo_loss}
\begin{aligned}
    %  L_{topo}(f,g) = 
     \min \limits_{\gamma \in \Gamma} \sum_{p \in \dgm(f)}||p-\gamma(p)||^2= 
    \sum_{p \in \dgm(f)}[\birth(p)-\birth(\gamma^*(p))]^2  
    +[\death(p)-\death(\gamma^*(p))]^2  
\end{aligned}
% \vspace{-10pt}
\end{equation}
where $\gamma^{*}$ is the optimal matching between two different point sets. 

Intuitively, this loss measures the minimal amount of necessary effort to modify the diagram of $\dgm(f)$ to $\dgm(g)$ by moving all dots toward their matches. Note there are more dots in $\dgm(f)$  (Fig.~\ref{fig:c}) than in $\dgm(g)$ (Fig.~\ref{fig:b});  there will usually be some noise in predicted likelihood map. If a dot cannot be matched, we match it to its projection on the diagonal line, $\{(d, b) | d = b\}$. This means we consider it as noise that should be removed. In this example, the extra connected component (a blue cross) in $\dgm(f)$ will be removed. For comparison, we also show in Fig.~\ref{fig:f} the matching between diagrams of the worse likelihood $f'$ and $g$. The cost of the matching is obviously higher, i.e., $L_{topo}(f',g) > L_{topo}(f,g)$. As a theoretical reassurance, it has been proven that this metric for diagrams is stable, and the loss function $L_{topo}(f,g)$ is Lipschitz with regard to the likelihood function $f$~\cite{cohen2007stability}. 

We state the following theorem regarding the correctness. The proof is relatively straightforward.
\begin{theorem}[Topological Correctness]
When the loss function $L_{topo}(f,g)$ is zero, the segmentation by thresholding $f$ at 0.5 has the same Betti number as $g$.
\end{theorem}
%\vskip -.15in
\vspace{-.05in}
%\proof
\myparagraph{Proof.} Assume $L_{topo}(f,g)$ is zero.
By Eq.~\eqref{topo_loss},  $\dgm(f)$ and $\dgm(g)$ are matched perfectly, \ie,  $p=\gamma^\ast(p), \forall p\in \dgm(f)$. The two diagrams are identical and have the same number of dots. 

Since $g$ is a binary-valued function, as we decrease the threshold $\alpha$ continuously, all topological
structures are created at $\alpha=1$. All of them except one are killed at $\alpha=0$. One remaining connected component will survive forever. The number of topological structures (Betti number) of $g^\alpha$ for any $0<\alpha<1$ is the same as the number of dots in $\dgm(g)$ plus 1. Note that for any $\alpha\in (0,1)$, $g^\alpha$ is the ground truth segmentation. Therefore, the Betti number of the ground truth is the number of dots in $\dgm(g)$. 
Similarly, for any $\alpha\in (0,1)$, the Betti number of $f^{\alpha}$ equals to the number of dots in $\dgm(f)$ plus 1. Since the two diagrams $\dgm(f)$ and $\dgm(g)$ are identical, the Betti number of the segmentation $f^{0.5}$ is the same as the ground truth segmentation. \footnote{Note that a more careful proof should be done for diagrams of different dimensions separately.}
\qed
% \subsection{Computing Topological Gradient}

% In this section, we derive the gradients of the topological loss, and try to incorporate the topological loss into neural networks.
% \begin{wrapfigure}{r}{5.5cm}
% \caption{A wrapped figure going nicely inside the text.}\label{wrap-fig:1}
% \includegraphics[width=5.5cm]{figure/Topo_fix.png}
% \end{wrapfigure} 

% \begin{wrapfigure}{r}{0.38\textwidth}
% \vspace{-30pt}
%   \begin{center}
%     \includegraphics[width=0.38\textwidth]{figure/Topo_fix.png}
%         \vspace{-30pt}
%   \end{center}

%   \caption{Birds}
% \vspace{-10pt}
% \end{wrapfigure}

\textbf{Topological gradient.} The loss function (Eq.~\eqref{topo_loss}) depends on crucial thresholds at which topological changes happen, e.g., birth and death times of different dots in the diagram. These crucial thresholds are uniquely determined by the locations at which the topological changes happen. When the underlying function $f$ is differentiable, these crucial locations are exactly \emph{critical points}, i.e., points with zero gradients. In the training context, our likelihood function $f$ is a piecewise-linear function controlled by the neural network predictions at pixels. For such $f$, a critical point is always a pixel, since topological changes always happen at pixels. Denote by $\omega$ the neural network parameters. For each dot $p \in \dgm(f)$ , we denote by $c_b(p)$ and $c_d(p)$ the birth and death critical points of the corresponding topological structure (See Fig.~\ref{fig:2c} for examples).

Formally, we can show that the gradient of the topological loss $\nabla_\omega L_{topo}(f,g)$  is:
%    \nabla_
\begin{equation}
% \vspace{-10p}
\label{gradient}
\begin{aligned}
%  \nabla_ \omega L_{topo}(f,g,\omega) = 
 \sum_{p \in \dgm(f)}2[f(c_b(p))-\birth(\gamma^*(p))]\frac{\partial f(c_b(p))}{\partial \omega}   
    +2[f(c_d(p))-\death(\gamma^*(p))]\frac{\partial f(c_d(p))}{\partial \omega}  
\end{aligned}
% \vspace{-10pt}
\end{equation}
To see this, within a sufficiently small neighborhood of $f$, any other piecewise linear function will have the same super level set filtration as $f$. The critical points of each persistent dot in $\dgm(f)$ remains constant within such small neighborhood. So does the optimal mapping $\gamma^*$. Therefore, the gradient can be straightforwardly computed based on the chain rule, as Eq.~\eqref{gradient}. When function values at different vertices are the same, the gradient does not exist. However, these cases constitute a measure zero subspace in the space of likelihood functions. In summary, $L_{topo}(f,g)$ is a piecewise differentiable loss function over the space of all possible likelihood functions $f$. 

\myparagraph{Intuition.} During training, we take the negative gradient direction, \ie,$-\nabla_ \omega L_{topo}(f,g)$. For each topological structure the gradient descent step is pushing the corresponding dot $p \in \dgm(f)$ toward its match $\gamma^*(p) \in \dgm(g)$. These coordinates are the function values of the critical points $c_b(p)$ and $c_d(p)$. They are both moved closer to the matched persistent dot in $\dgm(g)$. We also show the negative gradient force in the landscape view of function $f$ (blue arrow in Fig.~\ref{fig:2c}). Intuitively, force from the topological gradient will push the saddle points up so that the broken bridge gets connected.

% \begin{wraptable}{r}{5cm}
% \vspace{-10pt}
%   \centering
%   \begin{tabular}{|c|c|c|c|}
%     \hline
%     $i$ & $NW_{i1}$ & $NW_{i2}$ & $NW_{i3}$ \\\hline\hline
%     0 & 0 & 0 & 0 \\\hline
%     1 & 0 & 0 & 1 \\\hline

%   \end{tabular}
%   \caption{Some table}
% \end{wraptable}

\vspace{-5pt}
\subsection{Training a Neural Network}
\label{details}
\vspace{-5pt}
We present some crucial details of our training algorithm.
Although our method is architecture-agnostic, we select one architecture inspired by DIVE~\cite{fakhry2016deep}, which was designed for neuron image segmentation tasks. Our network contains six trainable weight layers, four convolutional layers and two fully connected layers. The first, second and fourth convolutional layers are followed by a single max pooling layer of size $2 \times 2$ and stride 2 by the end of the layer. Particularly, because of the computational complexity, we use a patch size of $65 \times 65$ during all the training process.

We use small patches ($65 \times 65$) instead of big patches/whole image. The reason is two-folds. First, the computational complexity of topological information is relative high. The computational complexity of persistent homology is $O(n^3)$, where $n$ is the patch size. Second, the matching process between the persistence diagrams of predicted likelihood map and ground truth can be quite difficult. For example, if the patch size is too big, there will be many persistent dots in Fig.~\ref{fig:b} and even more dots in Fig.~\ref{fig:e}. The matching process in Fig.~\ref{fig:c} is too complex and prone to  errors. \emph{By focusing on smaller patches, we localize topological structures and fix them one by one.}

% \begin{wrapfigure}{r}{0.1\linewidth}[h]
% \centering
% \rule{0.9\linewidth}{0.1\linewidth}[figure/Relative.pdf]
% % \caption{Dummy figure.}
% \label{fig:myfig}
% \end{wrapfigure}

\begin{wrapfigure}{r}{0.16\textwidth}
\centering 
\vspace{-5pt}
%  \begin{center}
    %   \hspace{-0.1in}
    \includegraphics[width=0.14\textwidth]{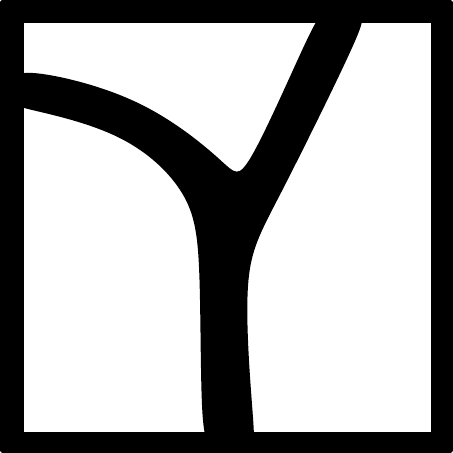}
        %  \vspace{-10pt}
%  \end{center}
%   \caption{Illustration of relative homology.}
  % Padding a frame to a patch will create two extra handles from a $Y$-shaped branch structure.}
%\vspace{-4pt}
\label{fig:padding}
\end{wrapfigure}
\textbf{Topology of small patches and relative homology.} 
The small patches ($65\times 65$) often only contain partial branching structures rather than closed loops. To have meaningful topological measure on these small patches, we apply \textit{relative persistent homology} as a more localized approach for the computation of topological structures. Particularly, for each patch, we consider the topological structures relative to the boundary. It is equivalent to padding a black frame to the boundary and compute the topology to avoid trivial topological structures. As shown in the figure on the right, with the additional frame, a $Y$-shaped branching structure cropped within the patch will create two handles and be captured by persistent homology. % See Fig.~\ref{fig:padding} for an illustration. 
Training using these localized topological loss can be very efficient via random patch sampling.

\textbf{Gradient of the topological loss.} 
%We want to emphasize the computation of gradient for each iteration during the training process. 
Eq.~\eqref{gradient} shows that the topological loss function is defined on \textit{critical pixels}. In each iteration, the predicted likelihood map from the neural network is different. Therefore, the topological structure and \textit{persistent dots} will also be quite different. The topological loss is computed based on different pixels (\textit{critical points}) at each iteration.

\vspace{-5pt}
\section{Experiments}
\vspace{-5pt}
We evaluate our method on six natural and biomedical datasets:
 \textbf{CREMI}\footnote{https://cremi.org/}, \textbf{ISBI12}~\cite{arganda2015crowdsourcing}, \textbf{ISBI13}~\cite{arganda20133d},
\textbf{CrackTree}~\cite{zou2012cracktree}, \textbf{Road}~\cite{mnih2013machine} and  \textbf{DRIVE}~\cite{staal2004ridge}. The first three are neuron image segmentation datasets. \textbf{CREMI} contains 125 images with resolution 1250*1250 with $4 \times 4 \times 40$nm. We use a three fold cross validation and report the mean performance over the validation set. Same setting applies to the other datasets. \textbf{ISBI12}~\cite{arganda2015crowdsourcing} contains 30 images. The resolution is 512*512 with $4 \times 4 \times 50$nm. \textbf{ISBI13}~\cite{arganda20133d} contains 100 images. The resolution is 1024*1024 with $3 \times 3 \times 30$nm.
These three datasets are neuron images (Electron Microscopy images). The task is to segment membranes and eventually partition the image into neuron regions. 
\textbf{CrackTree}~\cite{zou2012cracktree} contains 206 images of cracks in road. The resolution is 600*800. \textbf{Road}~\cite{mnih2013machine} has 1108 images from the Massachusetts Roads Dataset. The resolution is 1500*1500. \textbf{DRIVE}~\cite{staal2004ridge} is a retinal vessel segmentation dataset with 20 images. The resolution is 584*565.

% \subsection{Evaluation metrics used in this paper}
% \label{evaluation}

\begin{table*}[t]
% \begin{wraptable}{r}{9.8cm}%{0.8\textwidth}
\begin{center}
\caption{Quantitative results for different models on several medical datasets}
\label{table:medical}
\begin{tabular}{ccccccc}

Dataset & Method & Accuracy & ARI & VOI  & Betti Error\\
\hline
\multirow{5}{*}{\textbf{ISBI12}} &
\textbf{DIVE} & 0.9640  & 0.9434 & 1.235 & 3.187\\

~ & \textbf{U-Net} & \textbf{0.9678} & 0.9338 & 1.367 & 2.785\\
~ & \textbf{Mosin.} & 0.9532 & 0.9312 & 0.983 &1.238\\
~ & \textbf{TopoNet} & 0.9626 & \textbf{0.9444} & \textbf{0.782}  &\textbf{0.429} \\

\hline
\multirow{5}{*}{\textbf{ISBI13}} &
\textbf{DIVE} & \textbf{0.9642} & 0.6923 & 2.790 & 3.875\\

~ & \textbf{U-Net} & 0.9631 & 0.7031 &  2.583 &3.463\\
~ & \textbf{Mosin.} & 0.9578 & 0.7483 & 1.534 & 2.952\\
~ & \textbf{TopoNet} & 0.9569 & \textbf{0.8064} & \textbf{1.436}& \textbf{1.253}\\

\hline
\multirow{5}{*}{\textbf{CREMI}} &
\textbf{DIVE} & \textbf{0.9498} & 0.6532 & 2.513  & 4.378\\
~ & \textbf{U-Net} & 0.9468 & 0.6723 & 2.346 & 3.016\\
~ & \textbf{Mosin.} & 0.9467 & 0.7853 & 1.623 & 1.973\\
~ & \textbf{TopoNet} & 0.9456 & \textbf{0.8083} & \textbf{1.462}  & \textbf{1.113}\\

\hline

\end{tabular}
\end{center}
\vspace{-15pt}
% \end{wraptable}
\end{table*}

\textbf{Evaluation metrics.} We use four different evaluation metrics. \textbf{Pixel-wise accuracy} is the percentage of correctly classified pixels. 
The remaining three metrics are more topology-relevant.
The most important one is \textbf{Betti number error}, which directly compare the topology (number of handles) between the segmentation and the ground truth\footnote{Note we focus on 1-dimensional topology in evaluation and training as they are more crucial in practice.}. 
We randomly sample patches over the segmentation and report the average absolute difference between  their Betti numbers and the corresponding ground truth patches.
Two more metrics are used to indirectly evaluate the topological correctness: \textbf{Adapted Rand Index (ARI)} and \textbf{Variation of Information (VOI)}. They are often used in neuron image segmentation to compare the partitioning of the image induced by the segmentation. \textbf{ARI} is defined as the maximal F-score of the foreground-restricted Rand index, a measure of similarity between two clusters. On this version of the Rand index we exclude the zero component of the original labels (background pixels of the ground truth). \textbf{VOI} is a measure of the distance between two clusterings. It is closely related to mutual information; indeed, it is a simple linear expression involving the mutual information. 

% \ccinline{Move this to supp.}
% \textbf{Adapted Rand Index (ARI)} is defined as the maximal F-score of the foreground-restricted Rand index, a measure of similarity between two clusters. On this version of the Rand index we exclude the zero component of the original labels (background pixels of the ground truth). \textbf{VOI} is a measure of the distance between two clusterings. It is closely related to mutual information; indeed, it is a simple linear expression involving the mutual information. 

\textbf{Baselines.} \textbf{{DIVE}}~\cite{fakhry2016deep} is a state-of-the-art neural network that predicts the probability of every individual pixel in a given image being a membrane (border) pixel or not. \textbf{{U-Net}}~\cite{ronneberger2015u} is a popular image segmentation method trained with cross-entropy loss. \textbf{{Mosin.}}~\cite{mosinska2018beyond} uses the response of selected filters from a pretrained CNN to construct the topology aware loss.

% \begin{wraptable}{r}{9.8cm}%{0.8\textwidth}
\begin{table*}[t]
% \vspace{-10pt}
\centering
\begin{center}
\caption{Quantitative results for different models on some other datasets}
\label{table:ensemble}
\begin{tabular}{ccccccc}

Dataset & Method & Accuracy & ARI & VOI & Betti Error\\
\hline
\multirow{5}{*}{\textbf{DRIVE}} &
\textbf{DIVE} & \textbf{0.9549} & 0.8407 & 1.936 & 3.276 \\

~ & \textbf{U-Net} & 0.9452 & 0.8343 &  1.975 & 3.643\\
~ & \textbf{Mosin.} & 0.9543 & 0.8870   & 1.167 & 2.784\\
 ~ & \textbf{TopoNet} & 0.9521 & \textbf{0.9024} & \textbf{1.083} & \textbf{1.076}\\

\hline
\multirow{5}{*}{\textbf{CrackTree}} &
\textbf{DIVE} & \textbf{0.9854}  & 0.8634 & 1.570   & 1.576\\

~ & \textbf{U-Net} & 0.9821 & 0.8749 & 1.625  & 1.785\\
~ & \textbf{Mosin.} & 0.9833 & 0.8897 & 1.113  & 1.045\\
~ & \textbf{TopoNet} & 0.9826 & \textbf{0.9291} & \textbf{0.997} & \textbf{0.672} \\

\hline
\multirow{5}{*}{\textbf{Road}} &
\textbf{DIVE} & 0.9734 & 0.8201 & 2.368  & 3.598\\

~ & \textbf{U-Net} & \textbf{0.9786} & 0.8189 & 2.249  & 3.439\\
~ & \textbf{Mosin.} & 0.9754 & 0.8456 & 1.457  & 2.781\\
~ & \textbf{TopoNet} & 0.9728  & \textbf{0.8671} & \textbf{1.234}  & \textbf{1.275}\\

\hline

\end{tabular}
\end{center}
\vspace{-15pt}
\end{table*}
% \end{wraptable}

% \textbf{Warpping Error:} The warping error between some candidate labeling $T$ and a reference labeling $L^{*}$ is the Hamming distance (or equivalently squared Euclidean distance) between $L^{*}$ and the 'best warping' of $L^{*}$ onto $T$ : $D(T||L^{*})=\min \limits_{L \triangleleft L^{*}} ||T-L||^2$

\begin{figure*}[t]
\centering 
% \subfigure{
% \includegraphics[width=0.15\textwidth]{figure/patch1_ori.png}}
% % \vspace{-10pt}
% \subfigure{
% \includegraphics[width=0.15\textwidth]{figure/patch1_gt_revised.png}}
% \subfigure{
% \includegraphics[width=0.15\textwidth]{figure/patch1_a1_th_revised.png}}
% \subfigure{
% \includegraphics[width=0.15\textwidth]{figure/patch1_a2_th_revised.png}}
% \subfigure{ 
% \includegraphics[width=0.15\textwidth]{figure/patch1_a3_th_revised.png}}
% \subfigure{
% \includegraphics[width=0.15\textwidth]{figure/patch1_our_th_revised1.png}}

\vspace{-10pt}
\subfigure{
\includegraphics[width=0.15\textwidth]{figure/patch2_ori.png}}
\subfigure{
\includegraphics[width=0.15\textwidth]{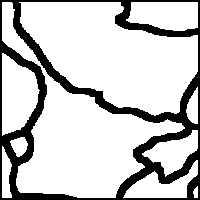}}
\subfigure{
\includegraphics[width=0.15\textwidth]{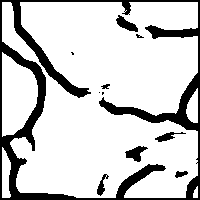}}
\subfigure{
\includegraphics[width=0.15\textwidth]{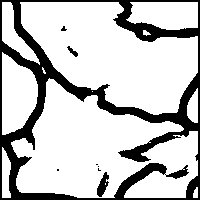}}
\subfigure{ 
\includegraphics[width=0.15\textwidth]{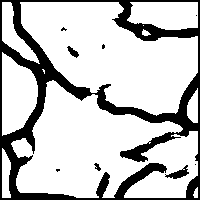}}
\subfigure{
\includegraphics[width=0.15\textwidth]{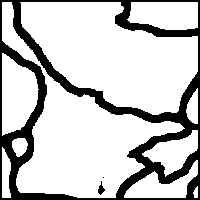}}

\vspace{-10pt}
\subfigure{
% \hspace{0.2in}
\includegraphics[width=0.15\textwidth]{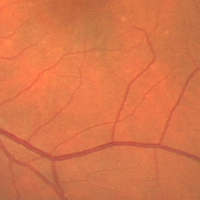}}
\subfigure{
\includegraphics[width=0.15\textwidth]{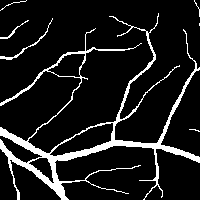}}
\subfigure{
\includegraphics[width=0.15\textwidth]{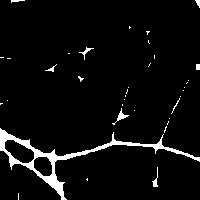}}
\subfigure{
\includegraphics[width=0.15\textwidth]{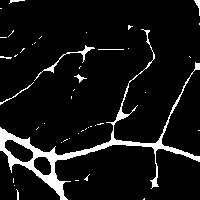}}
\subfigure{ 
\includegraphics[width=0.15\textwidth]{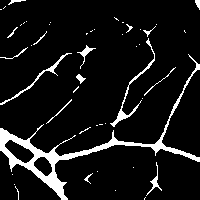}}
\subfigure{
\includegraphics[width=0.15\textwidth]{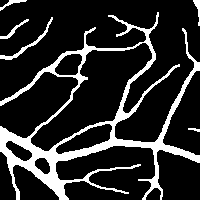}}

\vspace{-10pt}
\subfigure{
% \hspace{0.2in}
\includegraphics[width=0.15\textwidth]{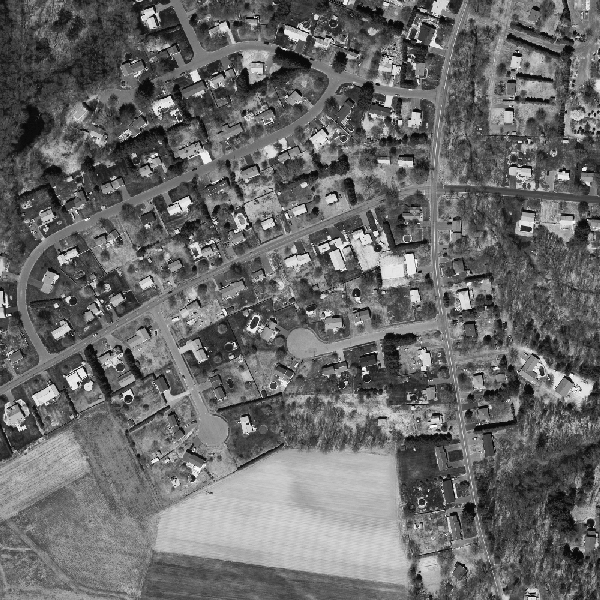}}
\subfigure{
\includegraphics[width=0.15\textwidth]{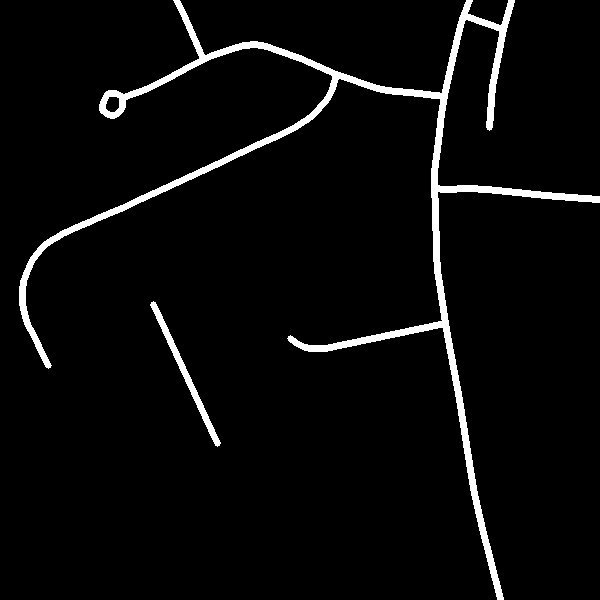}}
\subfigure{
\includegraphics[width=0.15\textwidth]{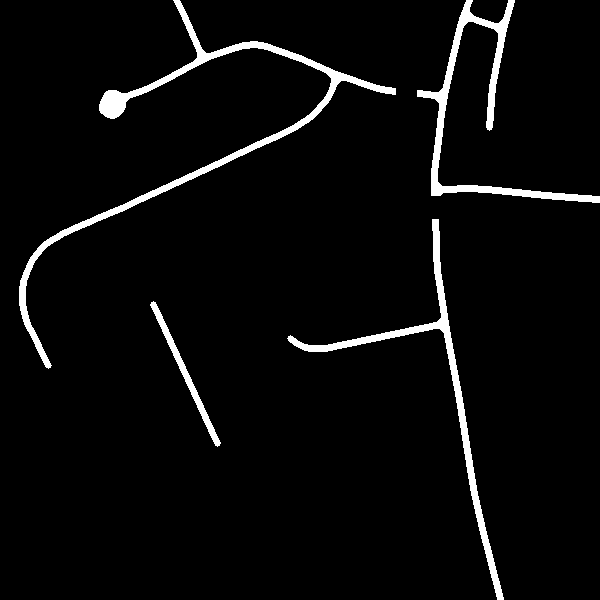}}
\subfigure{
\includegraphics[width=0.15\textwidth]{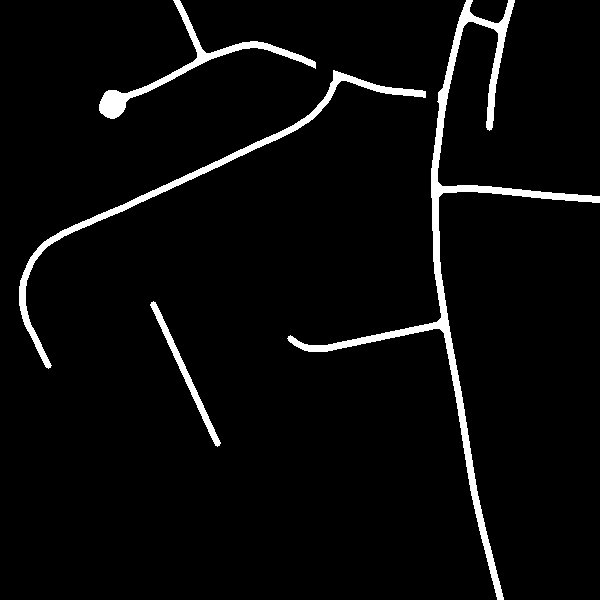}}
\subfigure{ 
\includegraphics[width=0.15\textwidth]{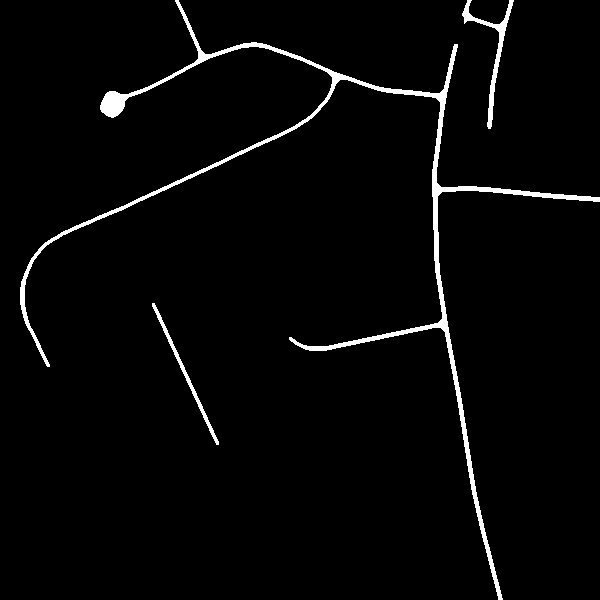}}
\subfigure{
\includegraphics[width=0.15\textwidth]{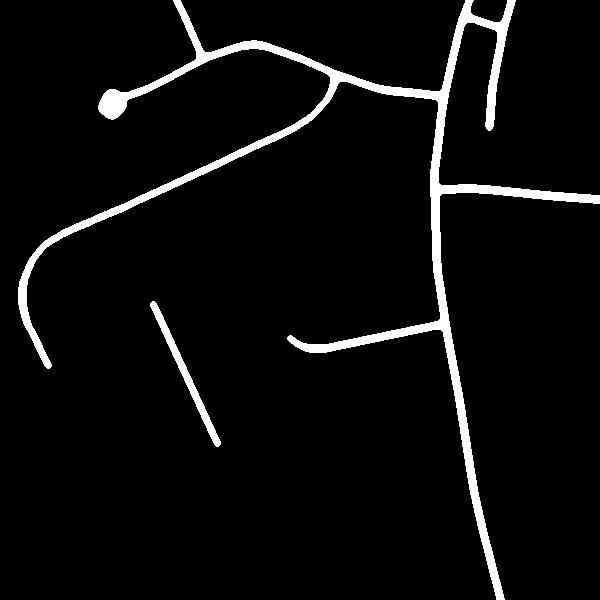}}

% \vspace{-10pt}
% \subfigure{
% % \hspace{0.2in}
% \includegraphics[width=0.15\textwidth]{figure/patch6_ori.png}}
% \subfigure{
% \includegraphics[width=0.15\textwidth]{figure/patch6_gt.png}}
% \subfigure{
% \includegraphics[width=0.15\textwidth]{figure/patch6_a1_th.png}}
% \subfigure{
% \includegraphics[width=0.15\textwidth]{figure/patch6_a2_th.png}}
% \subfigure{ 
% \includegraphics[width=0.15\textwidth]{figure/patch6_a3_th.png}}
% \subfigure{
% \includegraphics[width=0.15\textwidth]{figure/patch6_our_th.png}}

% \vspace{-10pt}
% \subfigure{
% % \hspace{0.2in}
% \includegraphics[width=0.15\textwidth]{figure/whole_ori.png}}
% \subfigure{
% \includegraphics[width=0.15\textwidth]{figure/whole_gt.png}}
% \subfigure{
% \includegraphics[width=0.15\textwidth]{figure/whole_a1.png}}
% \subfigure{
% \includegraphics[width=0.15\textwidth]{figure/whole_a2.png}}
% \subfigure{ 
% \includegraphics[width=0.15\textwidth]{figure/whole_a3.png}}
% \subfigure{
% \includegraphics[width=0.15\textwidth]{figure/whole_our_th.png}}

\vspace{-10pt}
\subfigure{
\hspace{.5pt}

\includegraphics[width=0.15\textwidth]{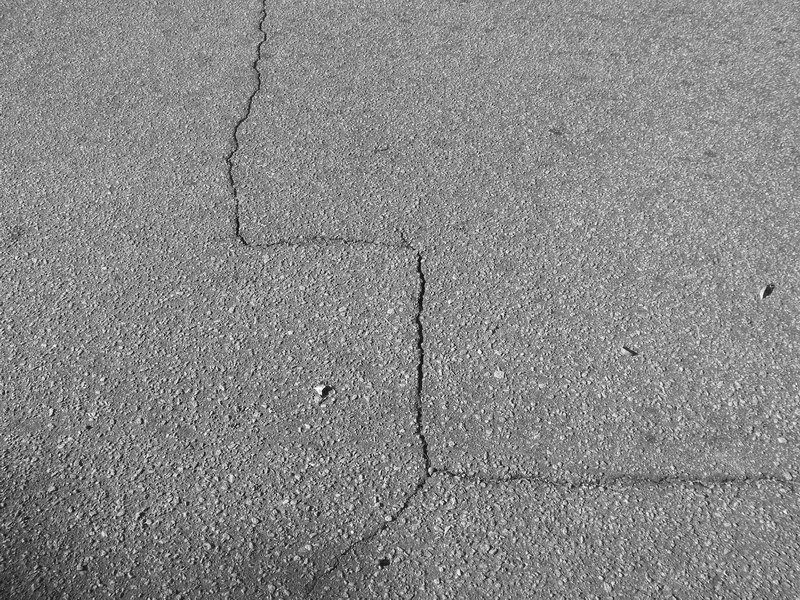}}
\subfigure{
% \hspace{-5pt}
\includegraphics[width=0.15\textwidth]{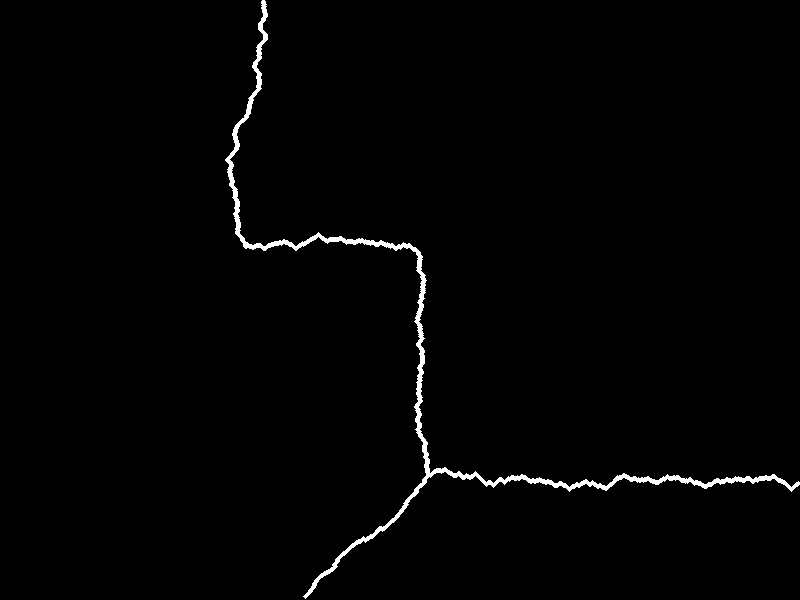}}
\subfigure{
\includegraphics[width=0.15\textwidth]{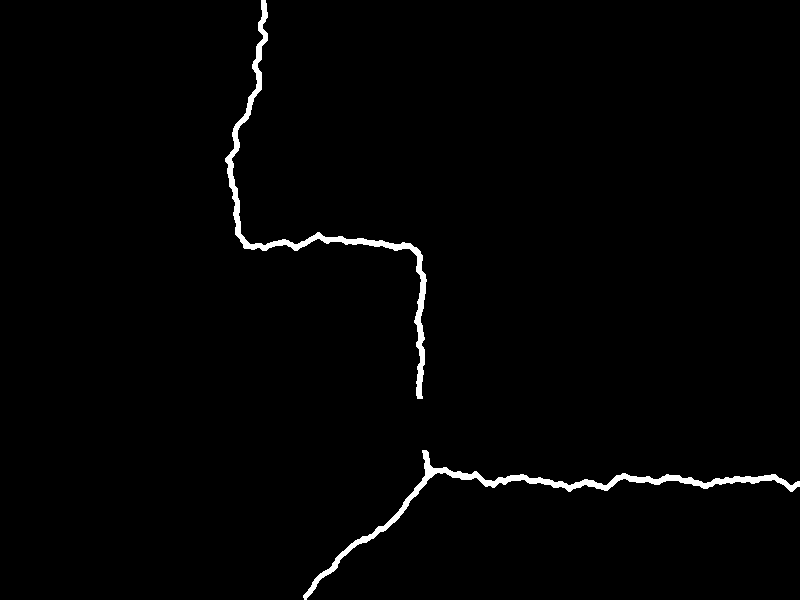}}
\subfigure{
\includegraphics[width=0.15\textwidth]{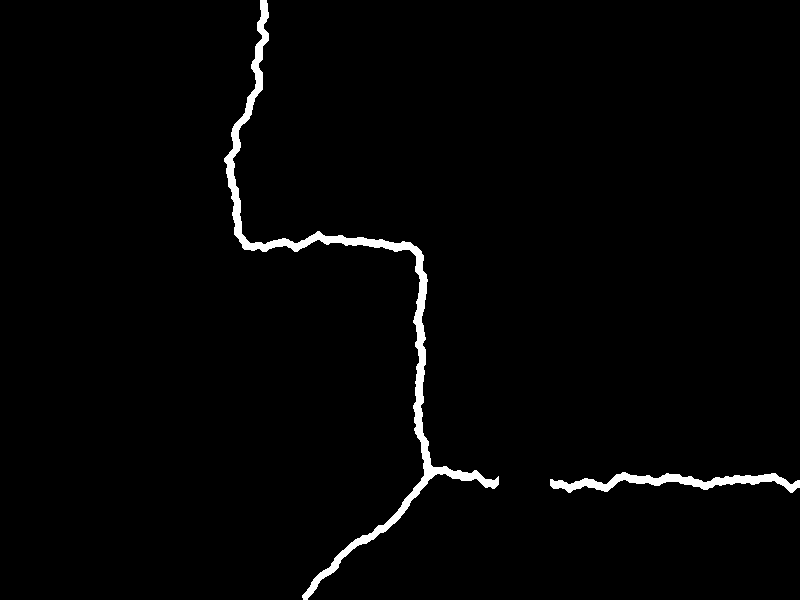}}
\subfigure{ 
\includegraphics[width=0.15\textwidth]{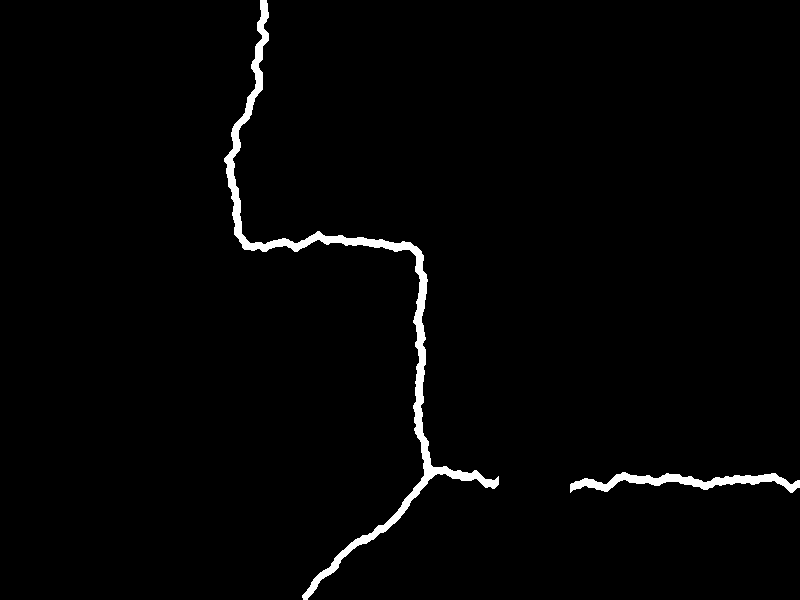}}
\subfigure{
% \hspace{5pt}
\includegraphics[width=0.15\textwidth]{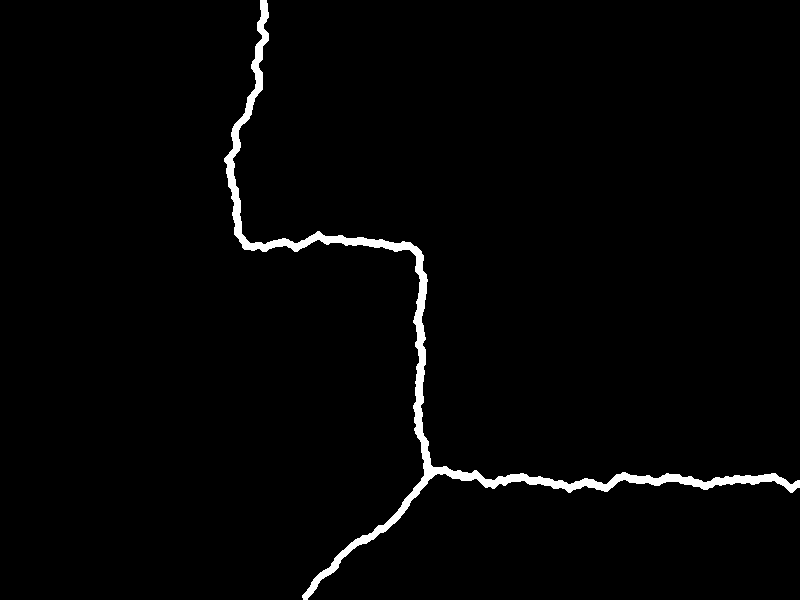}}
\vspace{-5pt}
\caption{Qualitative results of the proposed method compared to other models. From left to right, sample images, ground truth, results for \textbf{DIVE}, \textbf{U-Net}, \textbf{Mosin.} and \textbf{our proposed TopoNet}.}
\label{fig:result}
\vspace{-12pt}
\end{figure*}

\myparagraph{Quantitative and qualitative results.}
As stated above, we use pixel accuracy, and three topology-aware metrics: ARI, VOI and Betti Error. Tab.~\ref{table:medical} shows the quantitative results for three different neuron image datasets, ISBI12, ISBI13 and CREMI. Tab.~\ref{table:ensemble} shows the quantitative results for DRIVE, CrackTree and Road. We demonstrate that our method significantly outperforms existing methods in topological accuracy (in all three topology-aware metrics), without sacrificing pixel accuracy.
%
% \begin{wrapfigure}{r}{0.3\textwidth}
% % \vspace{-20pt}
%   \begin{center}
%     \includegraphics[width=0.4\textwidth]{figure/loss.eps}
%     % \hspace{-0.5in
%         %  \vspace{-10pt}
%   \end{center}
%
%   \caption{Loss VS Epoch.}
% % \vspace{-30pt}
% \label{fig:Loss}
% \end{wrapfigure}
%
Fig.~\ref{fig:result} shows qualitative results. TopoNet demonstrates more consistency in terms of structures and topology. It correctly segments fine structures such as membranes, roads and vessels, while all other methods fail to do so.  

% \begin{figure*}[ht]
%   \centering
%   \noindent\makebox[\textwidth][c] {
%     \includegraphics[width=0.5\paperwidth]{figure/result_analysis.png}}
%     \caption{Topology analysis: first line shows that some critical points are fixed and the second line shows hurting pixel accuracy.}
%       \label{fig:result}
% \end{figure*}

\begin{figure*}[t]
\centering 
\subfigure[]{\label{fig:complexity}
\includegraphics[width=0.35\textwidth]{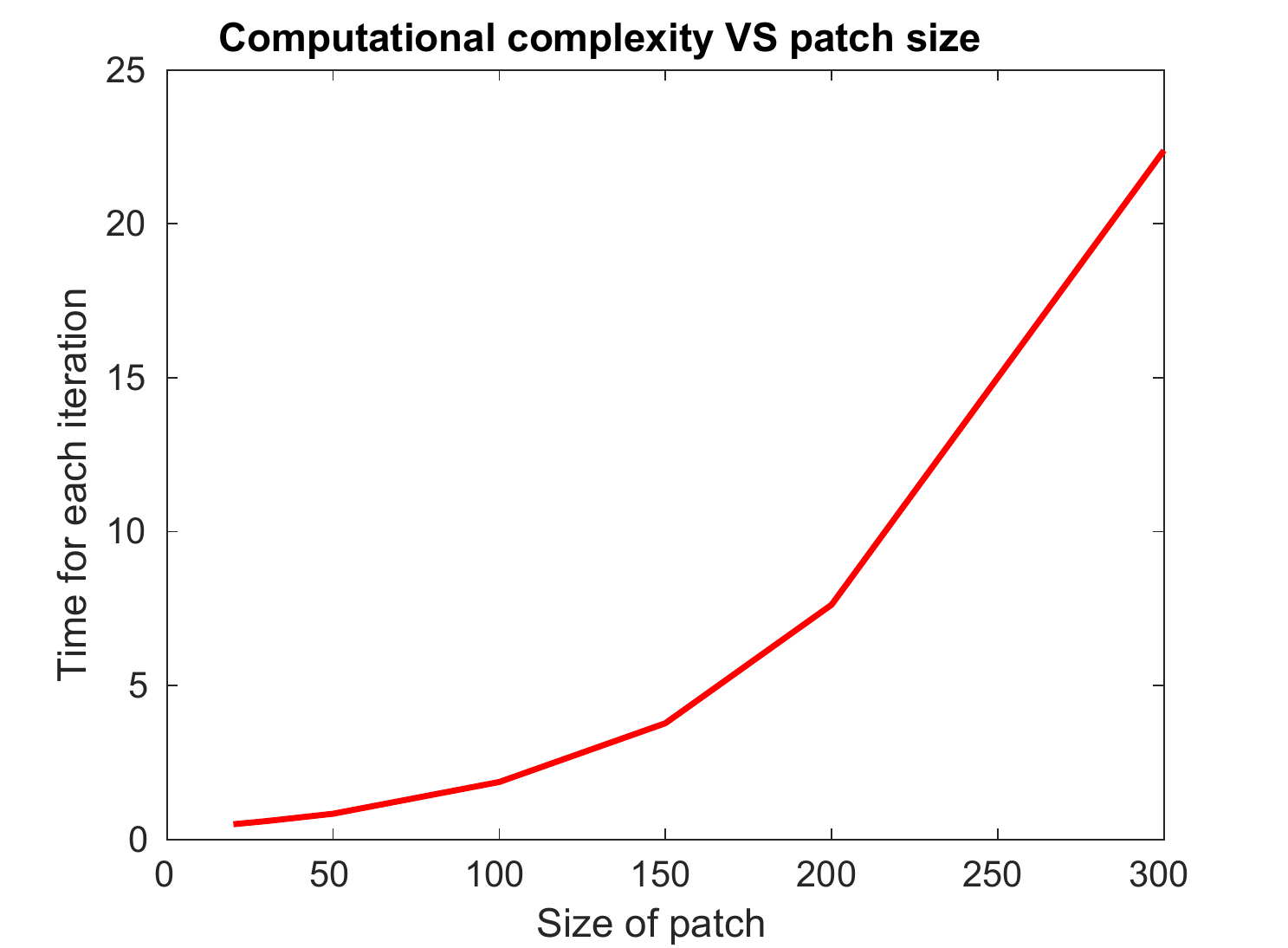}}
% \vspace{-10pt}
\subfigure[]{\label{fig:loss}
\includegraphics[width=0.35\textwidth]{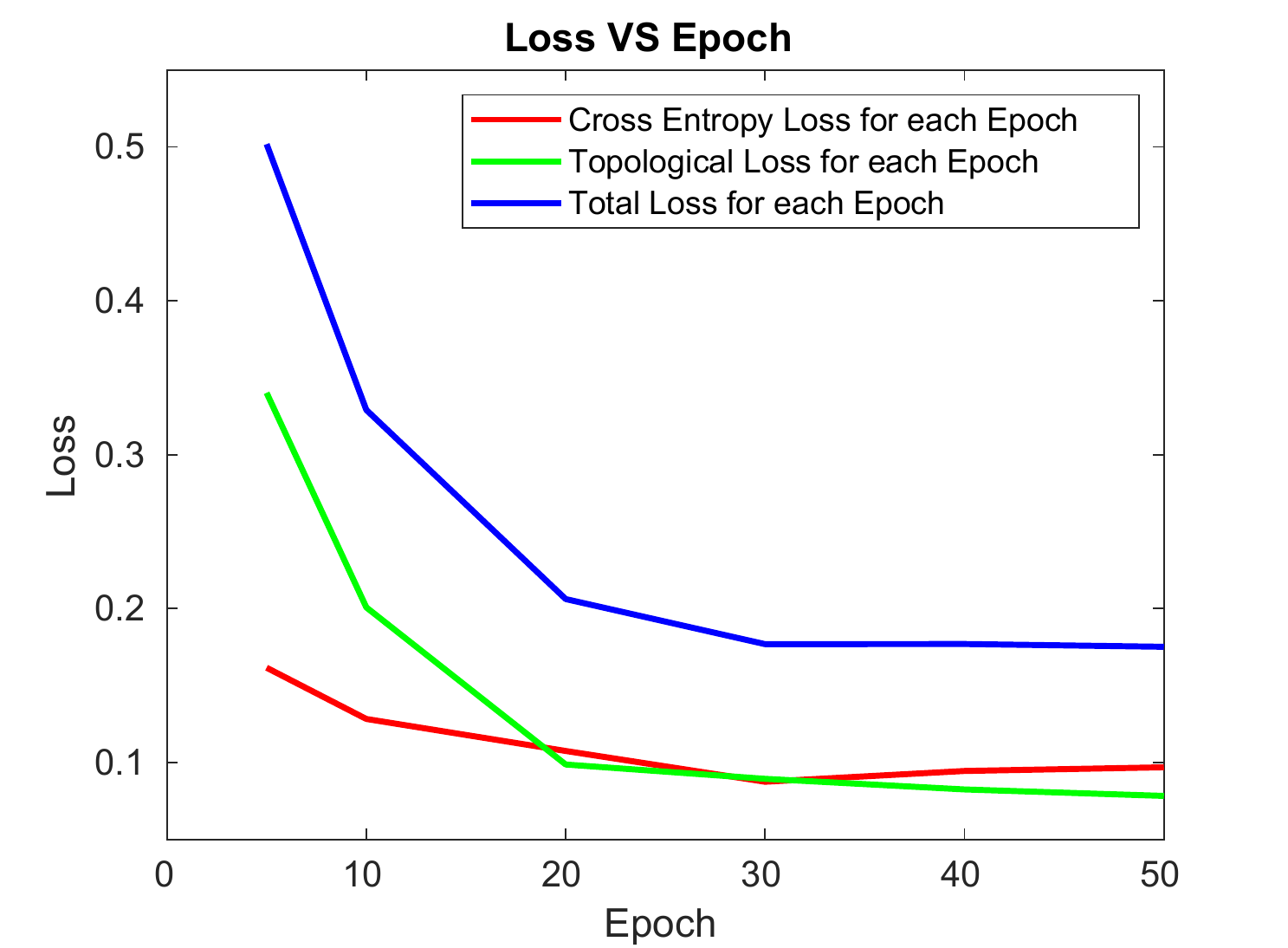}}
\vspace{-10pt}
\caption{\textbf{(a)} Topology computation complexity in terms of patch size. \textbf{(b)} Cross Entropy loss, Topological loss and total loss in terms of training epochs.}
\label{fig:ablation}
\vspace{-12pt}
\end{figure*}

\myparagraph{Computational time vs.~patch size.}
When incorporating our topological loss into deep neural networks, we test our algorithm with different patch sizes. During the training process, for each predicted likelihood map, we are supposed to compute its persistence diagram and the corresponding ground truth persistence diagram. The bigger the patch size is, the longer time it will take for this computation. The matching process will also be more complex and prone to errors (Sec.~\ref{details}). The empirical computation time in terms of patch size is illustrated in Fig.~\ref{fig:complexity}. To balance the training time and performance of the model, finally we choose $65 \times 65$ for all our experiments.

% \begin{wrapfigure}{r}{0.3\textwidth}
% \vspace{-20pt}
%   \begin{center}
%     \includegraphics[width=0.3\textwidth]{figure/complexity_size.eps}
%     \hspace{-0.3in}
%          \vspace{-10pt}
%   \end{center}

%   \caption{Computational complexity.}
% \vspace{-10pt}
% \label{fig:loss}
% \end{wrapfigure}

\myparagraph{Cross entropy and topological loss vs.~training epochs.}
In this part, we explore the trend of cross entropy loss and topological loss in terms of training epochs. The topological loss, cross entropy loss and total loss are illustrated in Fig.~\ref{fig:loss}. From the figure, we can find that after some epochs (in particular, after 30 epochs in Fig.~\ref{fig:loss}), the total loss becomes stable, while the cross entropy loss increases slightly as the topological loss decreases. This is reasonable; incorporating of topological loss may force the network to overtrain on certain locations (near critical pixels), and thus may hurt the overall pixel accuracy slightly. 
The pixel accuracy performance of our proposed method in Tab.~\ref{table:medical} and ~\ref{table:ensemble} also confirms this finding. 

\begin{figure*}[t!]
\centering 
\subfigure[]{\label{fig:gt}
\includegraphics[width=0.14\textwidth]{figure/patch2_gt_revised.png}}
\subfigure[]{\label{fig:epoch1}
\includegraphics[width=0.14\textwidth]{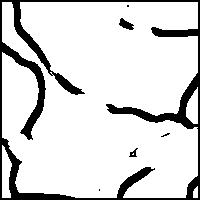}}
% \vspace{-10pt}
\subfigure[]{\label{fig:epoch2}
\includegraphics[width=0.14\textwidth]{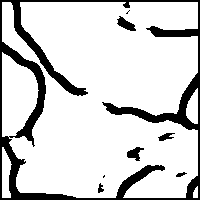}}
\subfigure[]{\label{fig:epoch3}
\includegraphics[width=0.14\textwidth]{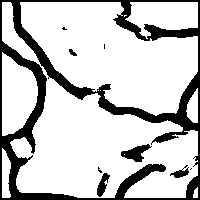}}
\subfigure[]{\label{fig:epoch4}
\includegraphics[width=0.14\textwidth]{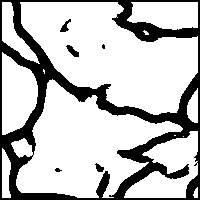}}
\subfigure[]{\label{fig:epoch5}
\includegraphics[width=0.14\textwidth]{figure/patch2_our_th_revised2.png}}
\vskip -0.05in
\caption{Segmentation at different training epochs. From left to right, ground truth, thresholded results after 10, 20, 30, 40 and 50 epochs.}
\label{fig:gradual}
\vspace{-12pt}
\end{figure*}

\myparagraph{The effectiveness of the topological loss}
We plot the thresholded results at different epochs to illustrate the effectiveness of our proposed topological loss (Fig.~\ref{fig:epoch1} to \ref{fig:epoch5}). As the training goes on, the fine structure of the membrane becomes more topologically consistent. This means that our topological term improves the quality of the predictions in terms of topology.
\vspace{-5pt}
\section{Conclusion}
\vspace{-5pt}
We introduce a new topological loss driven by persistent homology and incorporate it into end-to-end training of deep neural networks. Our method are particularly suitable for fine structure image segmentation. Quantitative and qualitative results show that the proposed topological loss term helps to achieve better performance in terms of topological-relevant metrics. Our proposed topological loss term is generic and can be incorporated into different architectures.

% \vfill
% \section*{References}
\bibliography{nips2019_arxiv} 
\bibliographystyle{plain}

\end{document}